# Transition to Adulthood for Young People with Intellectual or Developmental Disabilities: Emotion Detection and Topic Modeling

Yan Liu[1][0000−0003−3778−1325], Maria Laricheva[2], Chiyu Zhang[2], Patrick Boutet[2], Guanyu Chen[2], Terence Tracey[2], Giuseppe Carenini[2] and Richard Young[2]

[1] Carleton University, Ottawa ON, Canada
[2] The University of British Columbia, Vancouver BC, Canada
yanz.liu@carleton.ca

**Abstract.** Transition to Adulthood is an essential life stage for many families. The prior research has shown that young people with intellectual or development disabilities (IDD) have more challenges than their peers. This study is to explore how to use natural language processing (NLP) methods, especially unsupervised machine learning, to assist psychologists to analyze emotions and sentiments and to use topic modeling to identify common issues and challenges that young people with IDD and their families have. Additionally, the results were compared to those obtained from young people without IDD who were in transition to adulthood. The findings showed that NLP methods can be very useful for psychologists to analyze emotions, conduct cross-case analysis, and summarize key topics from conversational data. Our Python code is available at https://github.com/mlaricheva/emotion_topic_modeling.

**Keywords:** Transition to Adulthood, Intellectual or Development Disabilities, Emotion Detection, Sentiment Analysis, Topic Modeling, Natural Language Processing

## 1 Introduction

Transitioning to adulthood is an essential life stage for many families. It requires advanced planning and preparation, especially for families with a child who has intellectual or developmental disabilities (IDD). Youth with disabilities often experience more challenges than their peers when transitioning into adult roles, e.g., independent living, work, and engagement in social life [1,2]. Additionally, some services and supports come to an end when the person with IDD turns 18 years old, and parents need to identify new support resources for a growing adult.

Researchers have been trying to help young people with IDD to transition to adulthood in the past two decades [3]. Unfortunately, many young people with IDD still have been struggling in this process [4]. There is a need to understand more about what essential issues and concerns the youths and their families have and how their emotions are affected. Contextual action theory [5,6] has been used to study adult-hood transition and other career development issues. With a social constructionist approach, this framework helps people understand their own and others' thinking and behaviors in the joint actions. Because researchers need to understand actions in certain contexts, a pair of participants, also called dyads, are usually required to attend the study.

The conversational data obtained from this approach have been analyzed by psychologists using qualitative methods [4,7]. However, the data analysis process is complex. Pairs of researchers are usually assigned to conduct the initial analysis. The conversations are segmented minute by minute first, then analyzed line by line, and labeled according to a pre-established coding list. After the narrative summaries are developed from the initial analysis, the research team will review the analysis and summaries. One common strategy for psychologists is to conduct a cross-case analysis to understand the commonalities and differences among participants. It is usually conducted by grouping two or three cases at a time until the common themes and key information is extracted.

Recently, NLP methods, a combination of linguistics and machine learning techniques, have been developed rapidly for analyzing text data. The NLP methods have been shown to be a promising approach for analyzing conversational data [8,9]. Several studies have explored how NLP methods can be applied to large scale counselling conversational data. For example, Imel et al. [10] and Althoff et al. [11] analyzed the language of counselors to evaluate counsellor performance. Using online counseling conversations, Park et al. [12] analyzed client utterances to evaluate the therapy outcome.

However, the prior research focused on text counselling in mental health, so it is unknown whether NLP methods can work well for other psychological areas. Very few studies have investigated emotion issues and have not paid attention to cross-case analysis. Additionally, supervised machine learning with large scale data were used in these studies. However, many psychological studies have relatively small data as it is impossible to get large scale data due to limited funding and personnel. This study tries to bridge the gaps discussed here and explores how NLP methods, especially unsupervised machine learning methods, can be used for adulthood transition research.

## 2 Study Purpose

With a focus on adulthood transition research, especially for young people having IDD, this study aims to explore how to use NLP methods examine (1) common emotions and emotion intensity among all the dyads, (2) sentiment distributions across multiple transcripts, and (3) common topics and important issues discussed among all the dyads. Additionally, we compared the results to those obtained from young people without IDD to have a better understanding on emotions and essential issues for young people having IDD. More specifically, this study is to address three research questions:

RQ1: What common emotions can be identified from conversations made by all dyads whose family had a young person with IDD? Do the conversations made by young people without IDD have the same emotion distributions and intensities as compared to those made by families with a young person having IDD?

RQ2: How do dyads differ on their affects (positive vs. negative)?

RQ3: What common topics can be identified from conversations made by all dyads whose family had a young person with IDD? Do young people without IDD discuss the same topics as compared to those obtained from families with a young person having IDD?

## 3 Methods

### 3.1 Data Source

The present study is a secondary analysis of data that were collected from several studies related to adulthood transition between 2013 and 2016 [4, 13, 14]. The purposes of original studies were to understand joint actions and career goals of young people with IDD as well as those without IDD in their transition to adulthood. All dyads were invited to the lab for two meetings. Less than half of the dyads came back for the second meeting with 3-6 months apart. We pulled the conversations collected from both meetings for the final analysis because we found slight differences in the results of the two meetings. All the conversational data were labeled according to a pre-established coding list by researchers in the original studies.

Final dataset for the data analysis consisted of 63 text transcripts, with the duration about 10-20 minutes per transcript. The data included conversations between parents or a parent and an old sibling (29 transcripts), a parent and a child (9 transcripts), and two friends (25 transcripts). Friends, also called peers in this paper, were young people without IDD who were used as a reference to compare to young people having IDD.

### 3.2 Data Preprocessing

For emotion and sentiment analysis, we only used utterances related to emotions classified by the original studies, which is a small subset of the original data. A total of 1,552 utterances were included in our analysis. Most of original classifications were consistent with Plutchik's eight-emotion wheel, but some were more refined categories. To evaluate the results obtained from lexicon analysis, researchers pulled some original categories together to fit to the eight-emotion wheel. For example, "expresses gratitude", "expresses joy", "expresses like", and "express love" were all considered as "joy". We cleaned the data using the part-of-speech (POS) tagging provided in NLTK python package to remove digits, conjunctions, prepositions, modals, pronouns and particles. To simplify the work with the lexicon, an utterance was transformed to the list of words. However, we kept some words that would be considered as stop words in existing python packages because they expressed certain emotions. For example, word "aaaah" is associated with fear.

Different from emotion and sentiment analysis, all utterances were used for topic modeling analysis. In addition to the stop words provided by NLTK python package [15], we added 48 stop words. The words in the data were transformed to their numeric representations using term-frequency inverse-document-frequency (TF-IDF). Each transcript was split into sets with 25 utterances per set, which was identified to be an optimal document size by manual inspection of individual transcripts.

### 3.3 Analytical Methods

**Emotion Detection with NRC Lexicon**. We adopted National Research Council (NRC) lexicon [16] for emotion detection. Plutchik's eight-emotion wheel is utilized in NRC, including anger, anticipation, sadness, fear, trust, joy, surprise and disgust [17]. NRC lexicon provides a list of key words and their associations with eight emotions. Each word may be assigned to more than one emotion. For example, "money" has an association score of 0.586 with anticipation, 0.531 with joy, and 0.359 with trust. The association score can be used as an index of emotion intensity, i.e., a higher intensity score indicates a higher level of emotion.

We examined the distribution of emotions, i.e., the predicted proportion of each emotion, average emotion intensity, and the top common words related to each emotion. The average intensity score was obtained from the average of association scores from all utterances. When utterances were assigned to multiple emotions, the emotion with the highest association score was chosen for the emotion analysis.

To improve the quality of classification, we followed the recommendations by Mohammad and Turney [16] and adapted the lexicon to our study purpose. We removed the association between some keywords and emotions that we considered irrelevant. For instance, we removed the association between "kind" and joy because participants frequently treated it as adjective in a sentence, such as "It is kind of …"

**Sentiment Analysis with Bing Lexicon**. We used the Bing lexicon [18] to compare the positive and negative sentiment across 24 transcripts. This demonstration is to show readers that sentiment analysis can be used for cross-case analysis.

**Topic modeling**. Non-Negative Matrix Factorization (NMF) method was used for the topic modeling analysis and was shown to perform better than other methods when analyzing short documents [19]. NMF topic modeling decomposes the higher dimensional vectors of document-word representation, i.e., TF-IDF in this study, into a topic-document matrix with lower dimensional vectors and a word-topic matrix. We also used the combination of L1 and L2 norms for regularization in the NMF analysis.

## 4 Results

### 4.1 Emotion Detection with NRC Lexicon

Table 1 shows the predicted distribution of emotions using NRC lexicon, which was compared to the original classification made by researchers. Three types of emotions, i.e., *anticipation*, *joy*, and *trust*, were identified by NRC with a proportion above 0.2 for the families with young people having IDD. Similarly, *joy* and *trust* were identified for peers. However, peers had a relatively higher proportion on *disgust*, but lower on *anticipation*.

The results were consistent to most of researchers' classification. However, *trust* (0.277) was identified using NRC lexicon, but researchers did not include this class. It should be noted that *apprehension* (0.294), which is a subclass of *fear* based on the eight-emotion wheel, was included as an independent class by researchers because it was constructed by several psychological statuses (e.g., "expresses worry", "express confusion") that were semantically different from *fear*. We expected that the utterances originally classified as *apprehension* would be identified as *fear* by NRC. However, it was not shown in the NRC results.

**Table 1.** Distributions of Emotions Compared to Researchers' Classification (IDD vs. Peers)

|  | IDD | | Peers | |
| --- | --- | --- | --- | --- |
| emotion | Original | Lexicon | Original | Lexicon |
| apprehension | 0.294 | - | 0.204 | - |
| anger | 0.048 | 0.043 | 0.063 | 0.065 |
| **anticipation** | **0.237** | **0.204** | 0.098 | 0.163 |
| disgust | 0.081 | 0.023 | 0.151 | 0.058 |
| fear | 0.103 | 0.043 | 0.043 | 0.067 |
| **joy** | **0.201** | **0.308** | 0.349 | 0.288 |
| sadness | 0.014 | 0.084 | 0.024 | 0.076 |
| surprise | 0.022 | 0.018 | 0.068 | 0.037 |
| **trust** | - | **0.277** | - | **0.288** |

**Table 2.** The Average Emotion Intensity Scores across Time (IDD vs. Peers)

|  | IDD | | Peers | |
| --- | --- | --- | --- | --- |
| emotion | Time-1 | Time-2 | Time-1 | Time-2 |
| anger | 0.187 | 0.079 | 0.158 | 0.141 |
| **anticipation** | **0.654** | **0.579** | **0.377** | **0.325** |
| disgust | 0.105 | 0.059 | 0.114 | 0.145 |
| fear | 0.300 | 0.147 | 0.186 | 0.155 |
| **joy** | **0.690** | **0.678** | **0.537** | **0.452** |
| sadness | 0.309 | 0.237 | 0.171 | 0.159 |
| surprise | 0.163 | 0.145 | 0.156 | 0.095 |
| **trust** | **0.709** | **0.739** | **0.491** | **0.396** |

Table 2 shows the results of the average emotion intensity scores across two time points. Figure 1 visualizes the results and clearly shows that the overall emotion intensity levels for families with young people having IDD were much higher than their peers regardless the time changes. The top three emotions identified in the emotion distribution analysis, i.e., *anticipation, joy*, and *trust*, also showed high emotion inten-sity scores, all above 0.65 for families with young people having IDD. The same top emotions were identified for peers, but the intensity levels were lower, ranging from 0.325 to 0.537.

The emotions showed some changes over time for both populations although the changes were not in a large magnitude. For families with young people having IDD, the intensity level of positive emotions (*joy* & *trust*) was either remained around the same level or increased to a small degree over time, whereas the intensity level of negative emotions (*anger, disgust, fear*, & *sadness*) decreased over time. On the contrary, the results for peers suggested that the intensity of negative emotions (*disgust*) were increased slightly over time, whereas the intensity of positive emotions were decreased to some degree, *joy* (from 0.537 to 0.452) and *trust* (from 0.491 to 0.396).

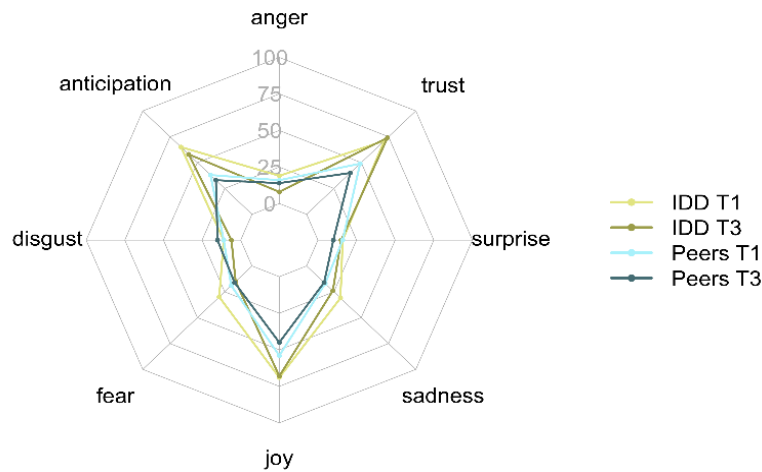

**Fig. 1.** The Average Emotion Intensity Scores across Time (IDD vs. Peers)

We also provided the top common words here to help researchers understand the connection of the emotion to the top common words used in the conversation. Table 3 provides a demonstration using data from families with young people having IDD, including *joy* and *trust*, and *fear*. We removed some words, e.g., "pretty", "good", because they are too generic. The results showed that "home, friends, money, love, family, happy, fun, hope, and independent" were related to people's *joy* and "school, money, love, happy, hope, doctor, friend, teacher, and mother" were associated with *trust*.

In general, the common words of positive emotions were more frequently appeared than those associated to negative emotions. The most frequently appeared common word related to *fear* is "worry". Additionally, "government" and "therapist" were shown in the conversations. It should be noted that "government" was more frequently appeared in the full transcripts, but it was only shown three times in this small subset of data. This common word reflected one common worry when parents talked about the government policy on the support to young people having IDD.

**Table 3.** Common Words Related to Emotions for Families with Young People Having IDD

| Fear | | Joy | | Trust | |
| --- | --- | --- | --- | --- | --- |
| word | count | word | count | word | count |
| worry | 25 | home | 41 | school | 37 |
| bad | 10 | friends | 21 | money | 20 |
| feeling | 7 | money | 20 | love | 16 |
| nervous | 6 | life | 18 | happy | 10 |
| hate | 5 | love | 16 | hope | 9 |
| concerned | 4 | special | 12 | doctor | 7 |
| scared | 3 | family | 11 | feeling | 7 |
| stealing | 3 | happy | 10 | friend | 6 |
| mad | 3 | fun | 9 | safe | 6 |
| government | 3 | hope | 9 | teacher | 6 |
| therapist | 3 | independent | 8 | mother | 5 |

### 4.2 Sentiment Analysis with Bing Lexicon

To demonstrate how sentiment analysis can be used for a cross-case analysis, we only randomly selected 24 transcripts from the conversations made by families with young people having IDD. Each transcript represents one case (two dyads). Figure 2 shows the distribution of the sentiment of the conversation on each transcript. The X-axis is the number of lines of utterances and each unit represents 20 lines; Y-axis is the frequency of utterances with positive sentiment above zero and negative below zero.

The sentiment was found to be different across cases. Some participants were more positive, some were more negative, and others had equally distributed positive and negative emotions. For example, case #7 was positive in the beginning and negative towards the end. Case #5 appeared to be negative over the whole conversation. Case #23 showed positive sentiment most of the time. This can be an easy tool for researchers to organize the transcripts before they conduct further in-depth qualitative analysis.

8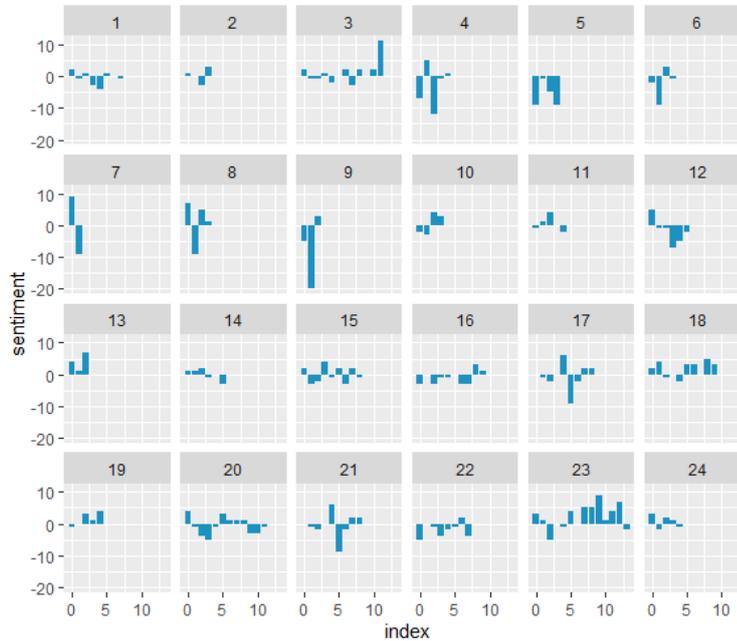

**Fig. 2.** Frequency distribution of utterances relative to positive and negative sentiment

### 4.3 Topic Modeling with Non-Negative Matrix Factorization (NMF)

Five main topics were identified by NMF from the conversations of families with young people having IDD (Figure 3). The final number of topics was chosen based on the best interpretation. These topics were also shown in the original studies [4, 20], including family, job search and work, financial issues (e.g., money management, eating, buying stuff), commuting and housing arrangement, and schooling. Additionally, the results showed that "independent" living may be a more concern for families with young people having IDD.

More topics were appeared in the conversations made by peers. Similarly, schooling (topic 4), family (topic 6), and financial issues (topic 7) were identified for peers. However, job interview was separated (topic 5) from work (topic 2); three new topics appeared, including functional communication (topic 1; e.g., help, feel, guess), communication for fun or social communication (topic 8), and driving car (topic 3).

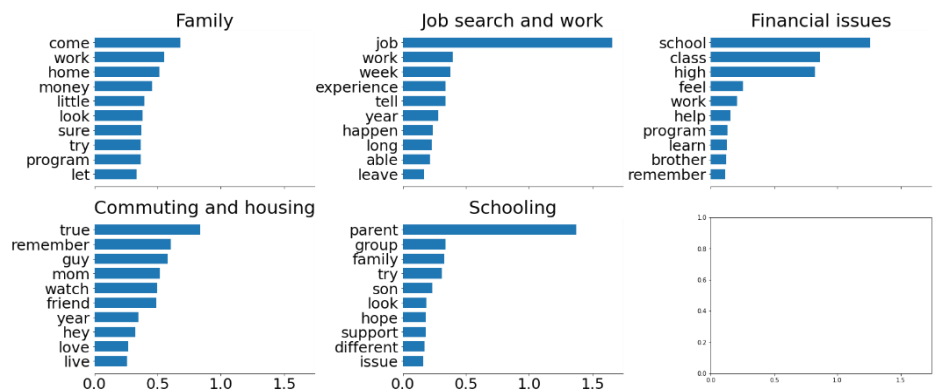

**Fig. 3.** Topic Modeling on Conversations of Families with Young People Having IDD

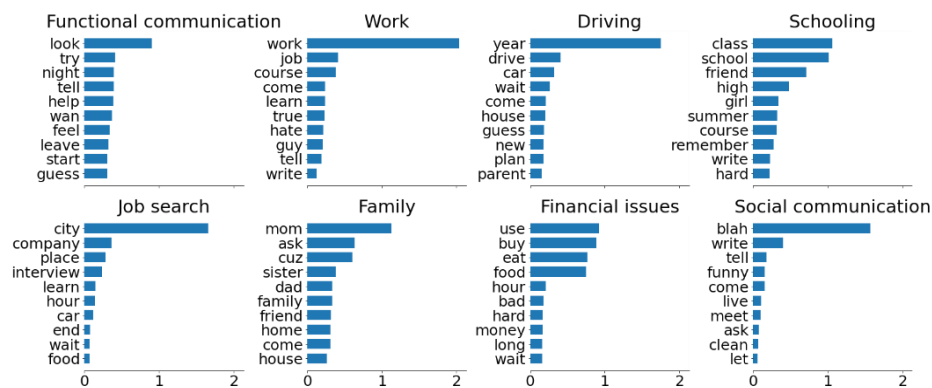

**Fig. 4**. Topic Modeling on Conversations of Young People without IDD

## 5      Conclusion and Discussion

This study investigated how NLP methods can be applied to adulthood transition research with relatively small conversational data. More specifically, this study explored how to use unsupervised machine learning methods to investigate emotions, sentiments, and topics in adulthood transition research, especially for families with youths having IDD. The results were also compared to young people without IDD.

The emotion and sentiment analysis have been neglected in earlier NLP applications. Our study showed that emotion detection and sentiment analysis can be reliable tools for psychologists to study participant emotions. Our results obtained from lexicon analysis were consistent with researchers' classification on emotions. Three top emotions were detected, including *joy, anticipation*, and *trust* for families with young people having IDD. Similar results were found for young people without IDD except *anticipation*. However, regardless of time changes, families with young people having

IDD showed an overall higher level of emotion intensity than young people without IDD.

Sentiment analysis can be an efficient and quick tool for cross-case analysis, which allows researchers to compare the participant affects across multiple transcripts. This may help researchers quickly categorize the transcripts for further qualitative analysis. Future studies are encouraged to make more use of this method.

Topic modeling identified five key topics for families with young people having IDD, including family, job search and work, financial issues, commuting and housing, and schooling, which echoed the findings from the original studies. Three more topics were identified for young people without IDD, i.e., functional communication, communication for fun, and driving. This may reflect the difference in essential issues between young people with and without IDD. Topic modeling is an efficient tool to exact key topics. However, this method may bring limitations to psychologist who want to extract deeper meaning from the conversation.

In general, our study has shown that NLP methods are useful for adulthood transition research and can be a practical tool for psychologists. Unsupervised machine learning methods work well with relatively small sample data. More research is encouraged to explore how NLP methods can be used for other research topic areas in psychology.

## References

1. Forte, M., Jahoda, A., Dagnan, D.: An anxious time? Exploring the nature of worries experienced by young people with a mild to moderate intellectual disability as they make the transition to adulthood: An anxious time? Br. J. Clin. Psychol. 50, 398–411 (2011).
2. Cronin, M.E.: Life Skills Curricula for Students with Learning Disabilities: A Review of the Literature. J. Learn. Disabil. 29, 53–68 (1996).
3. Kingsnorth, S., Healy, H., Macarthur, C.: Preparing for Adulthood: A Systematic Review of Life Skill Programs for Youth with Physical Disabilities. J. Adolesc. Health. 41, 323–332 (2007). https://doi.org/10.1016/j.jadohealth.2007.06.007
4. Marshall, S.K., Young, R.A., Stainton, T., Wall, J.M.: Transition to Adulthood as a Joint Parent-Youth Project for Young Persons With Intellectual and Developmental Disabilities. Intellect. Dev. Disabil. 56, 263–304 (2018). https://doi.org/10.1352/1934-9556-56.5.263
5. Young, R.A., Valach, L., Collin, A.: A contextual explanation of career. In: Brown, D. (ed.) Career choice and development. Jossey-Bass, San Francisco, CA (2002)
6. Young, R.A., Valach, L., Domene, J.F.: The action-project method in counseling psychology. J. Couns. Psychol. 52, 215–223 (2005). https://doi.org/10.1037/0022-0167.52.2.215
7. Valach, L., Young, R.A., Lynam, M.J.: Action Theory: A Primer for Applied Research in the Social Sciences. Greenwood Publishing Group (2002)
8. Carenini, G., Murray, G.: Methods for mining and summarizing text conversations. In: Proceedings of the 35th international ACM SIGIR conference on Research and devel-opment in information retrieval - SIGIR '12. p. 1178. ACM Press, Portland, Oregon, USA (2012)
9. Goldberg, Y.: Neural Network Methods for Natural Language Processing. Synth. Lect. Hum. Lang. Technol. 10, 1–309 (2017). https://doi.org/10.2200/S00762ED1V01Y201703HLT037
10. Imel, Z.E., Steyvers, M., Atkins, D.C.: Computational psychotherapy research: Scaling up the evaluation of patient–provider interactions. Psychotherapy. 52, 19–30 (2015). https://doi.org/10.1037/a0036841
11. Althoff, T., Clark, K., Leskovec, J.: Large-scale analysis of counseling conversations: An application of natural language processing to mental health. Trans. Assoc. Comput. Linguist. 4, 463 (2016)
12. Park, S., Kim, D., Oh, A.: Conversation Model Fine-Tuning for Classifying Client Utterances in Counseling Dialogues. Pap. Present. 2019 Annu. Conf. North Am. Chapter Assoc. Comput. Linguist. (2019)
13. Young, R.A., Marshall, S.K., Wilson, L.J., Green, A.R., Klubben, L., Parada, F., Polak, E.L., Socholotiuk, K., Zhu, M.: Transition to Adulthood as a Peer Project. Emerg. Adulthood. 3, 166–178 (2015). https://doi.org/10.1177/2167696814559304
14. Young, R.A., Marshall, S.K., Stainton, T., Wall, J.M., Curle, D., Zhu, M., Munro, D., Murray, J., Bouhali, A.E., Parada, F., Zaidman-Zait, A.: The transition to adulthood of young adults with IDD: Parents' joint projects. J. Appl. Res. Intellect. Disabil. 31, 224–233 (2017). https://doi.org/10.1111/jar.12395
15. Bird, S., Klein, E., Loper, E.: Natural language processing with Python, Analyzing text with the natural language toolkit. O'Reilly Media, Beijing (2009)
16. Mohammad, S.M., Turney, P.D.: Crowdsourcing a word-emotion association lexicon. Comput. Intell. 29, 436–465 (2013). https://doi.org/10.1111/j.1467-8640.2012.00460.x
17. PLUTCHIK, R.: Chapter 1 - A GENERAL PSYCHOEVOLUTIONARY THEORY OF EMOTION. In: Theories of Emotion. pp. 3–33. Elsevier Inc (1980)
18. Liu, B.: Sentiment analysis and opinion mining: Synthesis lectures on human language technologies. Morgan & Claypool (2012)
19. Lee, D.D., Seung, H.S.: Learning the parts of objects by non-negative matrix factorization. Nature. 401, 788–791 (1999). https://doi.org/10.1038/44565
20. Young, R.A., Marshall, S.K., Wilson, L.J., Green, A.R., Klubben, L., Parada, F., Polak, E.L., Socholotiuk, K., Zhu, M.: Transition to Adulthood as a Peer Project. Emerg. Adulthood. 3, 166–178 (2015). https://doi.org/10.1177/2167696814559304